\documentclass[ngerman,english]{bvm} 

\bvmdef\articlenumber{3057}

\bvmdef\type{V}

\date{}

\title{Learning the Update Operator for 2D/3D Image Registration}
\titlerunning{Learning the Update Operator for 2D/3D Image Registration}

\author{Srikrishna~Jaganathan$^{1,2}$, Jian~Wang$^{2}$, Anja~Borsdorf$^{2}$, Andreas~Maier$^{1}$}

\authorrunning{Jaganathan et al.}

\institute{$^{1}$Pattern~Recognition~Lab, FAU~Erlangen-Nürmberg, Erlangen, Germany.\\
$^{2}$Siemens~Healthineers~AG, Forchheim, Germany.}

\email{srikrishna.jaganathan@fau.de}

\begin{document}

%
\selectlanguage{english}

\maketitle

\begin{abstract}

Image guidance in minimally invasive interventions is usually provided using live 2D X-ray imaging. To enhance the information available during the intervention, the preoperative volume can be overlaid over the 2D images using 2D/3D image registration.  Recently, deep learning-based 2D/3D registration methods have shown promising results by improving computational efficiency and robustness. However, there is still a gap in terms of registration accuracy compared to traditional optimization-based methods. We aim to address this gap by incorporating traditional methods in deep neural networks using known operator learning. As an initial step in this direction, we propose to learn the update step of an iterative 2D/3D registration framework based on the Point-to-Plane Correspondence model. We embed the Point-to-Plane Correspondence model as a known operator in our deep neural network and learn the update step for the iterative registration. We show an improvement of 1.8 times in terms of registration accuracy for the update step prediction compared to learning without the known operator.

\end{abstract}

\section{Introduction}

In minimally invasive interventions, live 2D \mbox{X-ray} imaging is prominent for providing image guidance. The information available from 2D imaging alone is limited and can be augmented by overlaying the preoperative 3D volume over the 2D images. To obtain this overlay, the 3D volume needs to be accurately positioned such that, the corresponding structures are aligned between the 2D image and 3D volume. The optimal positioning of the 3D volume is accomplished using 2D/3D registration. A 2D/3D registration aims to find an optimal 3D transformation such that the misalignment between the 2D image and the 3D volume is minimized. Traditionally, to find this optimal transformation, the 2D/3D registration problem is formulated as an optimization problem. Depending on the application, 2D/3D registration can be classified into different sub-classes like modality of the images, the number of 2D views available, and constraints on the estimated transformation. A complete overview of the different traditional 2D/3D registration techniques and its different sub-class are summarized in \cite{3057-01}. Recently, \mbox{end-to-end} \mbox{Deep Learning (DL)-based} solutions have also been proposed for both single-view \cite{3057-02,3057-03} and \mbox{multi-view}  2D/3D registration \cite{3057-04} which shows significant improvement in terms of robustness and computational efficiency but often suffer in terms of registration accuracy.

In this work, we address \mbox{single-view} rigid 2D/3D registration between preoperative CT volume $\mathbf{V}$ and live fluoroscopic \mbox{X-ray} images $I_{\mathrm{Flr}}$. Generally, in single-view 2D/3D registration, the 3D CT volume is rendered using Digitally Reconstructed Radiograph (DRR) to obtain a simulated \mbox{X-ray} image $I_{\mathrm{DRR}}$. A similarity measure is defined to find the correspondences between the $I_{\mathrm{DRR}}$  and $I_{\mathrm{Flr}}$. Using this similarity measure, classical techniques model the 2D/3D registration as an optimization problem to find the optimal transformation such that the registration error is minimized. However, in \mbox{single-view} 2D/3D registration, due to the aperture problem, the 2D misalignment between the structures (obtained by finding 2D correspondences) gives only the observable 2D motion error. To find the 3D misalignment (thus the optimal 3D transformation), the unobservable motions in 2D should also be accounted for. This can be effectively constrained using the \mbox{Point-to-Plane Correspondence} (PPC) model~\cite{3057-05}. With the PPC model, a dynamic 2D/3D registration framework was proposed in \cite{3057-05}, which performs iterative 2D/3D registration by solving the PPC model at each iteration. It achieves \mbox{state-of-the-art} performance in terms of both registration accuracy and robustness. However, the framework relies on having accurate 2D correspondences between $I_{\mathrm{Flr}}$ and $I_{\mathrm{DRR}}$, which makes it sensitive to outliers in the estimated 2D correspondences.  To make the framework robust against outliers, a DL-based attention model was proposed~\cite{3057-06}. 

In this work, we extend the \mbox{PPC-based} registration framework to use a learned update operator. The update operator consists of 2D matching, weighing of the matches, and estimating the 3D rigid transformation such that it satisfies the PPC constraint. 
Since the PPC model is differentiable, it can be directly used as a known operator \cite{3057-07} and can be embedded as a layer in Deep Neural Network (DNN). Learning with the PPC model as a known operator has shown promising results when it was applied to learn correspondence weights for the 2D matches \cite{3057-06}. 
We propose to learn all the three steps of the PPC update operator fused into a single DNN, contrary to the previous attempts which only learned parts of the update operator~\cite{3057-06,3057-08}.

\section{Methods}

\subsection{PPC-based 2D/3D registration framework}

The \mbox{PPC-based} registration framework proposed in \cite{3057-05} is an iterative registration scheme and depends on the PPC constraint to estimate the optimal transformation at each iteration. Registration is performed between 3D CT volume $\mathbf{V}$ and 2D X-ray image $I_{\mathrm{Flr}}$ which are provided as inputs to the framework. Along with it, an initial transformation $\mathbf{T}_{\mathrm{init}}$ is required which provides a rough initial alignment.

The framework consists of an initialization step where the surface points along with their gradients are extracted from $\mathbf{V}$ using a 3D Canny edge detector, and the gradient of X-ray image ($\nabla I_{\mathrm{Flr}}$) is computed from $I_{\mathrm{Flr}}$. The initialization step is performed only once, and the values are cached. 
After the initialization step, the update operation is performed iteratively until convergence. The update operation consists of the following steps.
Based on the current transformation estimate, the gradient image of DRR ($\nabla I_{\mathrm{DRR}}$) is rendered. Surface points with gradients perpendicular to the current viewing direction are selected as contour points $\mathbf{w}_i$ where $i \in {1,\dots, N}$ with $N$ contour points. The contour points $\mathbf{w}_i$ are projected into  $\nabla I_{\mathrm{DRR}}$ to get the projected contour points $\mathbf{p}_i$. Now, 2D matching is performed to find the corresponding projected contour points $\mathbf{p}'_i$ in $\nabla I_{\mathrm{Flr}}$. The correspondence set ($\mathbf{p}_i,\mathbf{p}'_i$) along with contour points $\mathbf{w}_i$, and its gradients $\mathbf{g}_i$ are used to compute the 3D motion update based on the PPC model. If the 2D matches are noisy, a weighted version of the PPC model is used to reduce the effect of the noisy matches. The weighted PPC model gives a linear constraint $\mathbf{W} \mathbf{A} \mathbf{dv} = diag(\mathbf{W}) \mathbf{b}$, where $\mathbf{A}$ and  $\mathbf{b}$ are the data terms which are computed based on the estimated 2D correspondences. $\mathbf{W}$ is a diagonal matrix which consists of weights for each correspondences. Each correspondence contributes one row of $\mathbf{A},\mathbf{b}$ and $\mathbf{W}$ . Closed form solution $\mathbf{dv} =(\mathbf{WA}^+) (diag(\mathbf{W})\mathbf{b})$ using the pseudo inverse of $\mathbf{A}$ is used for computing the 3D motion. The rigid 3D transformation matrix $\mathbf{T}_\mathrm{est}$  is computed from the 3D motion vector $\mathbf{dv}$, which serves as an input for the next update step.

\subsection{Learnable update operator for PPC-based 2D/3D registration framework}

\begin{figure}[h]
    \centering
    \includegraphics[width=0.9\linewidth]{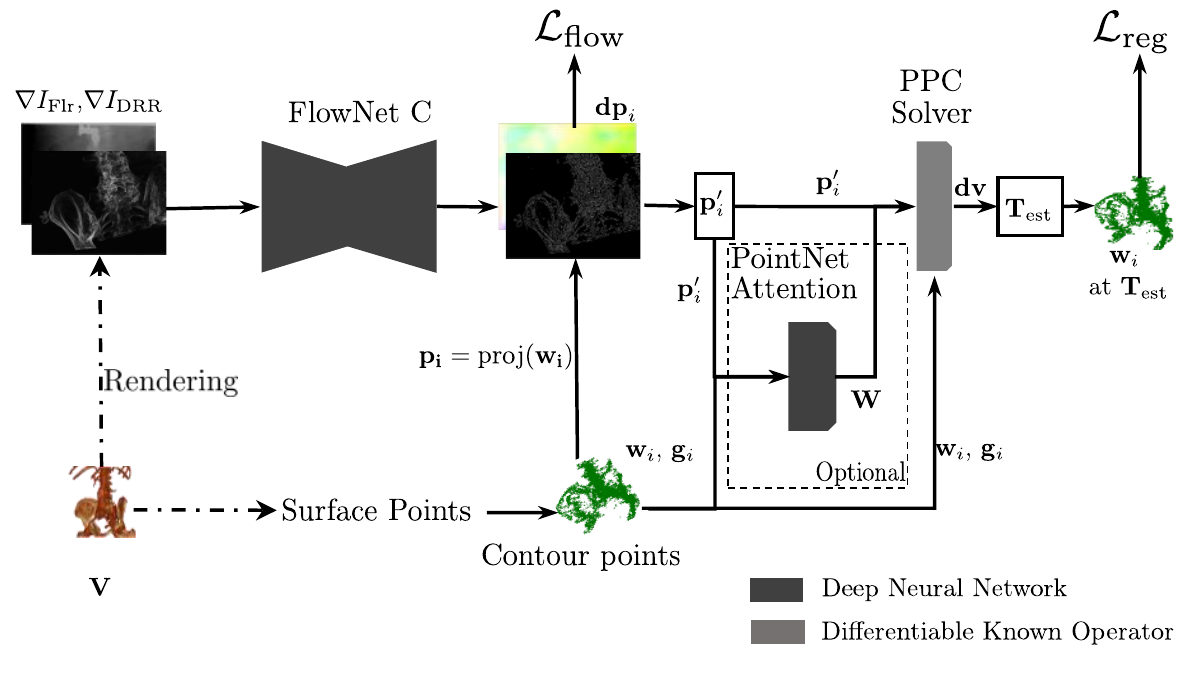}
    \caption{Schematic of the proposed update operator for PPC based iterative registration framework.}
    \label{3057-fig-01}
\end{figure}

The update operator in \mbox{PPC-based} registration framework finds $\mathbf{T}_\mathrm{est}$ given a set of inputs ($\nabla I_{\mathrm{Flr}}$,$\nabla I_{\mathrm{DRR}}$, $\mathbf{w}_i$, $\mathbf{g}_i$, $\mathbf{p}_i$). Fig.~\ref{3057-fig-01} shows the network architecture used for training the update operator. The FlowNet C architecture \cite{3057-09} is used for estimating 2D matches between $\nabla I_{\mathrm{Flr}}$ and $\nabla I_{\mathrm{DRR}}$ by predicting optical flow $\mathbf{dp}_i$ at projected contour points $\mathbf{p}_i$ similar to \cite{3057-08}. Using $\mathbf{p}_i'= \mathbf{p}_i + \mathbf{dp}_i$ and the 3D information $(\mathbf{w}_i,\mathbf{g}_i)$, correspondence weighting matrix $\mathbf{W}$ is predicted using PointNet Attention model similar to ~\cite{3057-06}. This step is optional and we define two architectures PPC Flow and PPC Flow Attention, where we use the PointNet Attention only for the later. The 3D motion $\mathbf{dv}$ is computed using the PPC Solver which gives us the current estimated 3D transformation $\mathbf{T}_\mathrm{est}$. Applying $\mathbf{T}_\mathrm{est}$, the contour points $\mathbf{w}_i$ are updated . We use the mean Target Registration Error (mTRE) as registration loss $\mathcal{L}_{\mathrm{reg}}$. It is computed using $ \frac{1}{N} \sum_{i}^{N} ||\mathbf{T}_\mathrm{est}(\mathbf{w}_i) - \mathbf{T}_\mathrm{gt}(\mathbf{w}_i)||$ for a set of N contour points and ground truth transformation $\mathbf{T}_\mathrm{gt}$. Additionally, we also use optical flow loss $\mathcal{L}_{\mathrm{flow}}$ by computing average End Point Error (EPE) at  projected contour points $\mathbf{p}_i$ as proposed in~\cite{3057-08}, to make the network training more stable.

\subsection{Experimental Setup}

The data set used for training and evaluation is from reconstruction data of Cone Beam CT (CBCT) for the vertebra region~\cite{3057-06}. The data set consists of 56 acquisitions from 55 patients. The data is split into 38 patients for training, 5 patients for validation, and 12 patient for testing. The training samples are created using random initial transformation from the ground truth registration using different viewing directions from the volume similar to \cite{3057-08}. We generate random initial transformations with an initial mTRE in the range of [0, 45] mm.
Each sample consists of ($\nabla I_{\mathrm{Flr}}$,$\nabla I_{\mathrm{DRR}}$, $\mathbf{w}_i$, $\mathbf{g}_i$, $\mathbf{p}_i$) along with $\mathbf{T}_{gt}$. There are about 18,000 such samples for training and validation, and 8,000 samples for testing.

The update operator of the \mbox{PPC-based} registration framework performs one registration iteration. We train and evaluate the performance of the update operator using three different models namely Flow, PPC Flow, and PPC Flow Attention.
The Flow and PPC Flow models are used to compare the effects of known operator learning. In the Flow model, only the 2D matching is learned by training the model with ground truth flow annotations at contour points $\mathbf{p}_i$ as proposed in \cite{3057-08}. Here, we use the PPC solver only for evaluating its registration performance after one update. PPC Flow uses the same optical flow architecture as the Flow model along with the PPC model embedded as a known operator.  
PPC Flow Attention integrates the PPC Flow with the attention model~\cite{3057-06}  to learn both correspondence estimation and weighing in an end-to-end manner. Both PPC Flow and PPC Flow Attention are trained using both registration loss (mTRE) and optical flow loss (average EPE).

We train all the networks for 200,000 iterations (100 epochs) with a batch size of 16. We use the ADAM optimizer with a learning rate of 1e-4 and weight decay is used as a regularizer with a decay rate of 1e-6.

For evaluation, we use mTRE after one registration update which measures the registration error. Lower values indicate better registration accuracy. In addition, we also use reduction factor which is computed for samples whose ${\mathrm{mTRE}_{\mathrm{i+1}}} <{\mathrm{mTRE}_{\mathrm{i}}} $  as  ($ 1 - \frac{\mathrm{mTRE}_{\mathrm{i+1}}}{\mathrm{mTRE}_{\mathrm{i}}} $) where $\mathrm{mTRE}_{\mathrm{i}}$ is the error before update step and $\mathrm{mTRE}_{\mathrm{i+1}}$ is the error after the application of the update step. It is set to 0.0 for samples whose error increases after the update. It takes a value between 0.0 to 1.0 where higher values are better. The reduction factor indicates how much of the error can be reduced with one update step. 

\section{Results}

\begin{figure}[htb]
	\setlength{\figbreite}{0.3\textwidth}
	\centering
	\subfigure[]{\includegraphics[width=\figbreite]{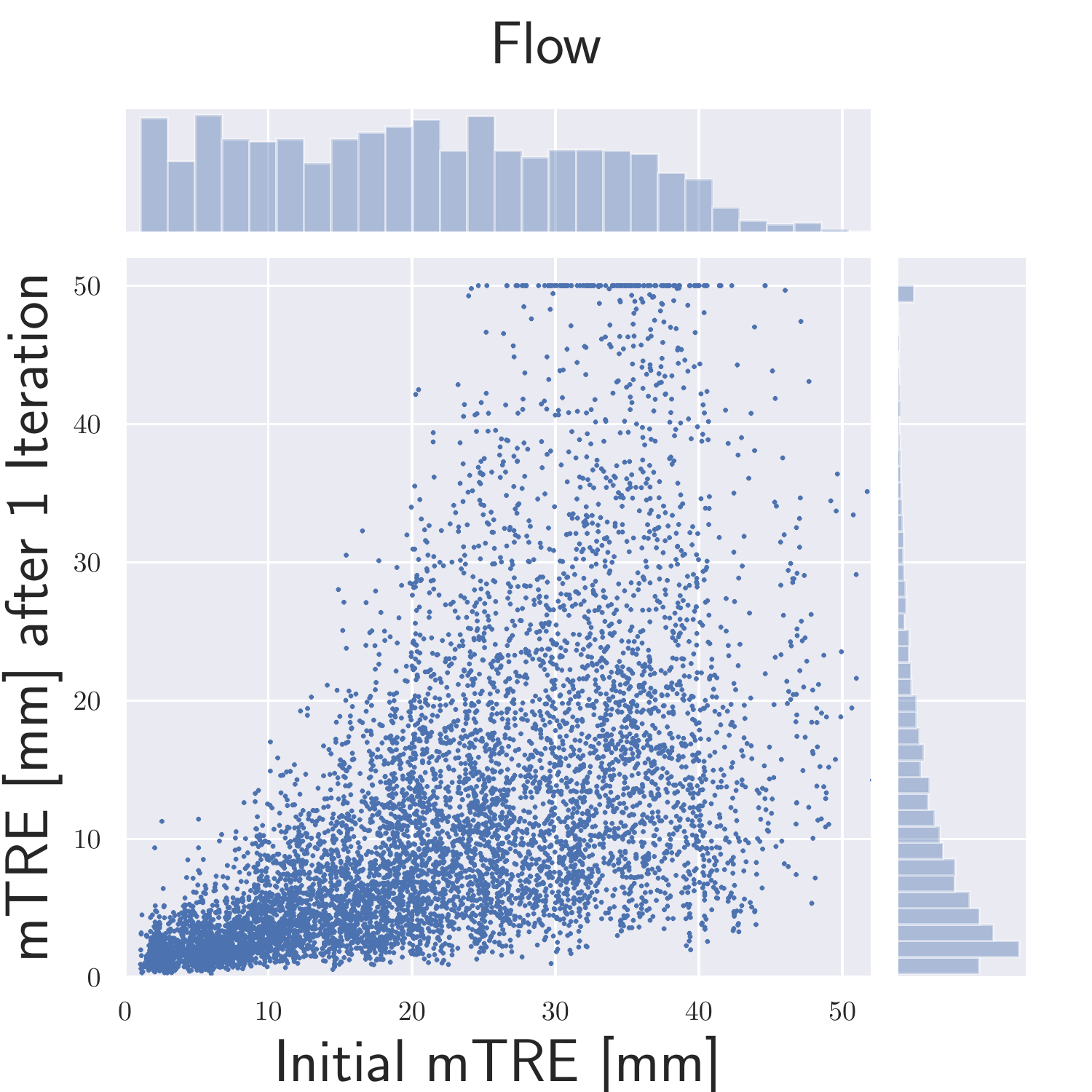}}
	\subfigure[]{\includegraphics[width=\figbreite]{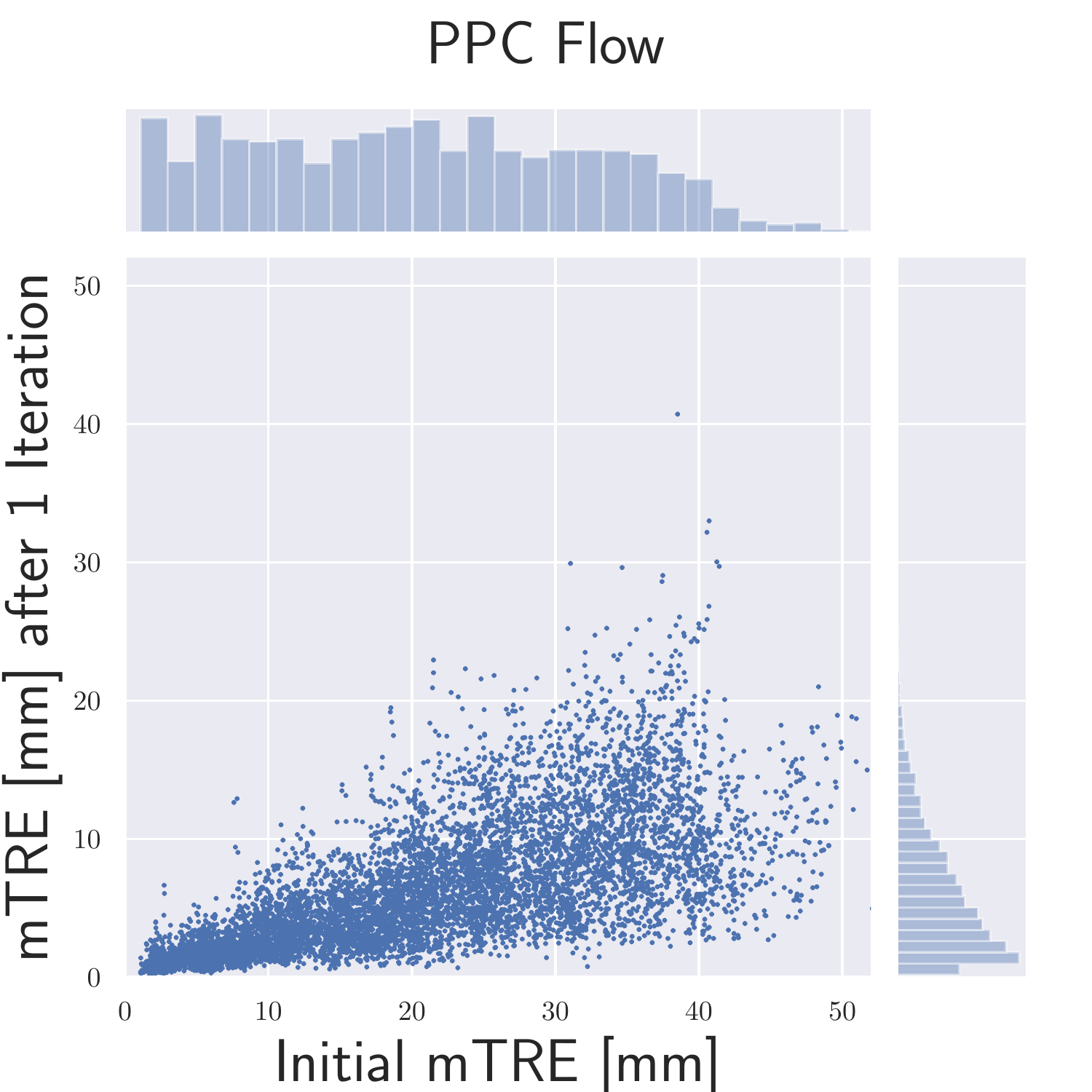}}
	\subfigure[]{\includegraphics[width=\figbreite]{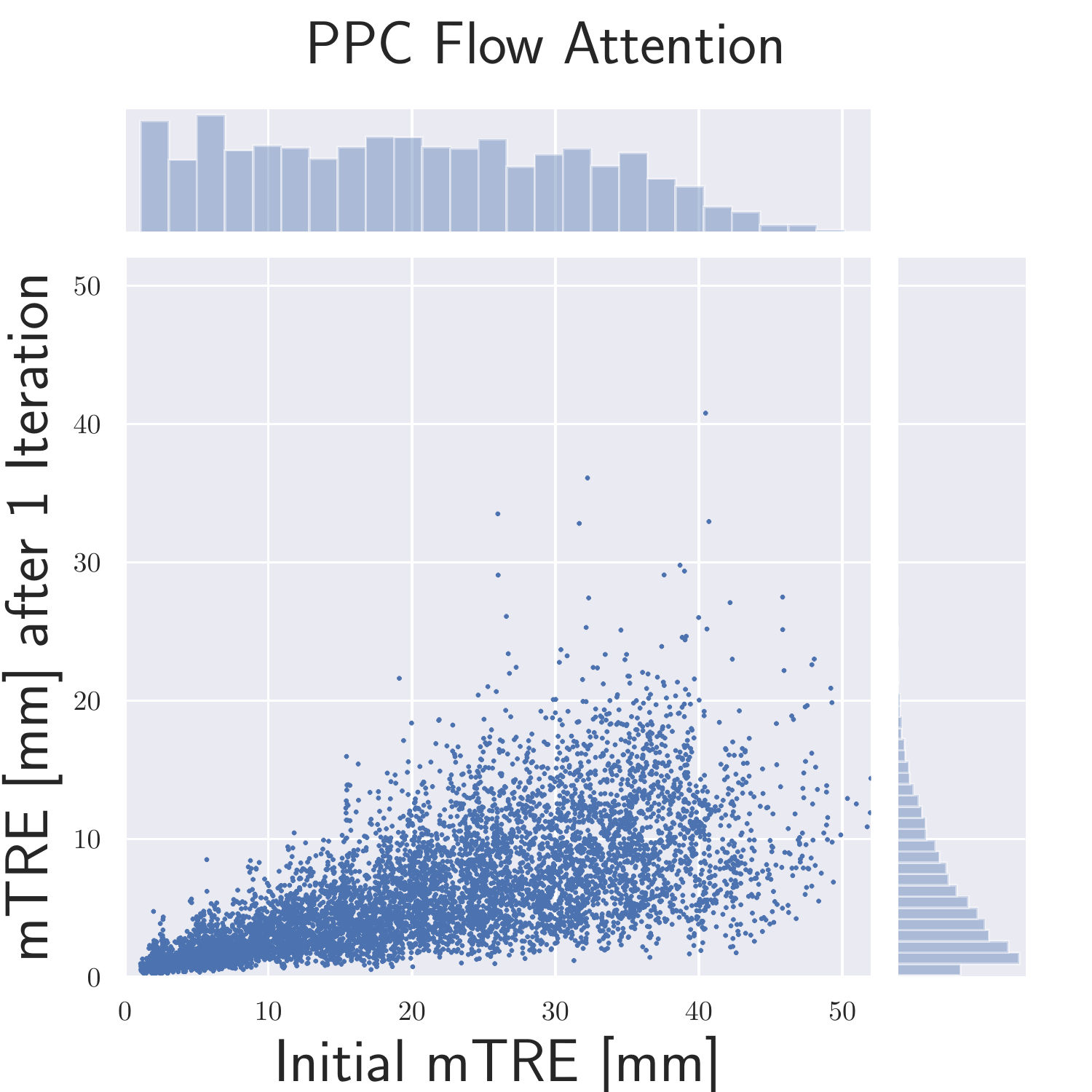}}

	\caption{Error distribution for the test data set, showing registration errors before and after one update operation for (a) Flow, (b) PPC Flow, and (c) PPC Flow Attention. For better visualization, the values of mTRE after 1 iteration are clipped at 50mm.}
	\label{3057-fig-02}
\end{figure}

 The performance of the different models evaluated on the test data set is summarized in Table~\ref{3057-tab-01}. The Flow, PPC Flow, and PPC Flow Attention models achieve a mTRE ($\mu \pm \sigma$ computed over all samples) of $11.27 \pm 11.66 $ mm, $6.21 \pm 4.70 $ mm and $5.88 \pm 4.50 $ mm respectively. The reduction factor for Flow, PPC Flow, and PPC Flow Attention is ($\mu \pm \sigma$ computed over all samples) $0.47 \pm 0.38 $, $0.68 \pm 0.17 $, $0.69 \pm 0.17$ respectively. 
 Fig.~\ref{3057-fig-02}  compares the change in registration error after one update for all the models used. 

\begin{table}[h]
\centering
\begin{tabular}{p{0.25\textwidth}p{0.10\textwidth}p{0.10\textwidth}p{0.10\textwidth}p{0.20\textwidth}p{0.20\textwidth}}
\hline 
 \multirow{2}{*}{} & \multicolumn{3}{l}{Percentile mTRE [mm]} & mTRE [mm] & Reduction Factor \\
  & $\displaystyle 50^{^{th}}$ & $\displaystyle 75^{^{th}}$ & $\displaystyle 95^{^{th}}$ & $\displaystyle \ \mu \ \pm \ \sigma $ & $\displaystyle \ \mu \ \pm \ \sigma $ \\
\hline 
Initial            & 20.18 & 30.09 & 39.75 &     20.59  $\pm$        11.76 &  \\
Flow               &  7.45 & 15.20 & 34.59 &     11.27 $\pm$        11.66 &     0.47 $\pm$       0.38 \\
PPC Flow           &  5.03 &  8.80 & 15.41 &      6.21 $\pm$        4.70 &     0.68 $\pm$       0.17 \\
PPC Flow Attention &  $\mathbf{4.70}$ &  $\mathbf{8.31}$ & $\mathbf{14.68}$ & $\mathbf{5.88}$ $\pm$       $\mathbf{4.50}$ &     $\mathbf{0.69}$ $\pm$       $\mathbf{0.17}$ \\
 \hline
\end{tabular}
\caption{Registration error measured using mTRE in [mm] of different networks after one update operation, evaluated on test data set. The initial mTRE indicates the initial registration error. The $\displaystyle 50^{th} ,75^{th} ,95^{th} $ percentile errors are provided to indicate percentage of samples having an error $\leq$ the indicated value. The mTRE ($ \mu \ \pm \sigma$) and the reduction factor ($\mu \ \pm \sigma$) summarizes the model performance over all the samples.}
\label{3057-tab-01}
\end{table}

\section{Discussion and Conclusion}

We proposed a \mbox{DL-based} extension to learn the update operation of the PPC-based iterative 2D/3D registration framework. PPC Flow learns the 2D matching operation with PPC model as a known operator. It improves the performance of the update operation by a factor of 1.8 compared to Flow model in terms of mean reduction in registration error as shown in Table~\ref{3057-tab-01}. Our intuition is that, the PPC model enforces constraints to learn correspondences that are relevant for the registration thus improving the performance. PPC Flow Attention learns 2D matching and weighting together in an end-to-end manner. This improves the performance over the PPC Flow model as it can effectively discard the outliers that might still be present in the correspondences predicted using PPC Flow.

The proposed update operator serves as an initial step in extending the \mbox{PPC-based} registration framework to a \mbox{DL-based} module which retains the strength of classical 2D/3D registration techniques while also providing all the benefits of \mbox{DL-based} methods. However, there are still some areas that need to be addressed for achieving this goal. Especially, in cases where we start with small registration errors, the learned update operator has minimal influence. We plan to address this issue in our future work either by unrolling the iterative registration or with multiple learned update operators. Additionally, one can explore how to make the update operator tailor-made for specific clinical applications.

\paragraph{}
{\bf Disclaimer}: The concepts and information presented in this paper are based
on research and are not commercially available.

\bibliographystyle{bvm}

\bibliography{3057}
\marginpar{\color{white}E\articlenumber} 
\end{document}